\documentclass[]{ceurart}
\usepackage{amsmath}
\usepackage[most]{tcolorbox}

\usepackage[utf8]{inputenc}

\usepackage{listings}
\usepackage{pgfplots}
\usepackage{pgfplotstable}
\pgfplotsset{compat=1.18}
\usepackage{xcolor}
\usepackage{tikz}
\usetikzlibrary{shapes.geometric, arrows.meta, positioning, calc}

\begin{document}
\copyrightyear{2025}
\copyrightclause{Copyright for this paper by its authors. Use permitted under Creative Commons License Attribution 4.0 International (CC BY 4.0).}
\conference{CLEF 2025 Working Notes, 9 -- 12 September 2025, Madrid, Spain}

\title{JU-NLP at Touch\'e: Covert Advertisement in Conversational AI-Generation and Detection Strategies }

\title[mode=sub]{Notebook for the Touch\'e Lab at CLEF 2025}

%%
%% The "author" command and its associated commands are used to define the authors and their affiliations.
\author{Arka Dutta}[%
  email=arka08652@gmail.com,
]
\fnmark[1]
\cormark[1]
\address[1]{Department of Computer Science and Engineering, Jadavpur University, Kolkata, 700032, India}
\author{Agrik Majumdar}[
  email=agrik.maz33@gmail.com
]
\fnmark[1]
\author{Sombrata Biswas}[
  email=sombrata.biswas@gmail.com
]
\fnmark[1]
\author{Dipankar Das}
\author{Sivaji Bandyopadhyay}
\cortext[1]{Corresponding author.}
\fntext[1]{These authors contributed equally.}

%%
%% The abstract is a short summary of the work to be presented in the article.
\begin{abstract}
  This paper proposes a comprehensive framework for the generation of covert advertisements within Conversational AI systems, along with robust techniques for their detection. It explores how subtle promotional content can be crafted within AI-generated responses and introduces methods to identify and mitigate such covert advertising strategies. For generation (Sub-Task~1), we propose a novel framework that leverages user context and query intent to produce contextually relevant advertisements. We employ advanced prompting strategies and curate paired training data to fine-tune a large language model (LLM) for enhanced stealthiness. For detection (Sub-Task~2), we explore two effective strategies: a fine-tuned CrossEncoder (\texttt{all-mpnet-base-v2}) for direct classification, and a prompt-based reformulation using a fine-tuned \texttt{DeBERTa-v3-base} model. Both approaches rely solely on the response text, ensuring practicality for real-world deployment. Experimental results show high effectiveness in both tasks, achieving a precision of 1.0 and recall of 0.71 for ad generation, and F1-scores ranging from 0.99 to 1.00 for ad detection. These results underscore the potential of our methods to balance persuasive communication with transparency in conversational AI.
\end{abstract}

\begin{keywords}
  llm finetuning\sep
  stealth advertisement\sep
  binary classification\sep
  sentence transformers\sep
  cross encoder\sep
  retrieval-augmented generation\sep
  context-aware generation\sep
  prompt-based learning\sep
  DeBERTa\sep
  transformer models
\end{keywords}

\maketitle

\section{Introduction}

The detection and generation of covert advertisements in conversational AI is an emerging challenge that intersects language understanding, marketing ethics, and human-computer interaction. As conversational agents and retrieval-augmented generation (RAG) systems become increasingly integrated into user-facing platforms, there is a growing concern around the insertion of native advertisements that may subtly influence user behavior without clear disclosure. The ability to generate such responses in a contextually relevant yet stealthy manner, as well as to accurately detect them post-generation, is essential for preserving trust and transparency in AI-mediated communication~\cite{lewis2021retrievalaugmentedgenerationknowledgeintensivenlp}.

The 2025 edition of the Touché shared task addresses this concern by introducing two sub-tasks~\cite{kiesel:2025b}:
\begin{itemize}
  \item Sub-Task 1: Given a user query and supporting document context, generate a relevant response that optionally includes a covert advertisement for a given item or service.
  \item Sub-Task 2: Given a response generated by a conversational system, classify whether it contains a native advertisement.
\end{itemize}

In our participation as Team JU-NLP, we propose tailored solutions for both sub-tasks. For Sub-Task~1, we construct a high-quality training dataset by leveraging a large language model (LLM) as a judge to evaluate a multiset of responses generated by a pretrained LLM across iterative prompts, scoring them based on advertisement detectability. These preference-labeled pairs are then used to fine-tune a LLM (like, Mistral-7B model~\cite{jiang2023mistral7b}) using the ORPO (Odds Ratio Preference Optimization) training framework~\cite{hong2024orpomonolithicpreferenceoptimization}, which encourages the generation of contextually coherent yet covert promotional content. For Sub-Task~2, we explore two complementary detection strategies: (1) a fine-tuned CrossEncoder (\texttt{all-mpnet-base-v2}) that performs binary classification using only the response text, and (2) a prompt-based reformulation of the task utilizing a fine-tuned \texttt{DeBERTa-v3-base} model, aimed at improving detection performance through instruction-style inputs and enhanced contextual understanding~\cite{he2023debertav3improvingdebertausing, song2020mpnetmaskedpermutedpretraining}.

This paper is structured as follows: Section~\ref{sec:gen} presents our method for advertisement generation using a preference-tuned LLM model. Section~\ref{sec:detection} describes the techniques employed for advertisement detection. Finally, Section~\ref{sec:conclusion} concludes the paper with a discussion of the results and potential directions for future work.

\section{Covert Ad Insertion in Conversational AI}\label{sec:gen}
In this work, we introduce a novel framework for generating covert advertisements that are seamlessly integrated into contextually relevant responses. The proposed system embeds promotional content related to a product or service in a manner that preserves the coherence and informativeness of the response while minimizing the likelihood of detection as an advertisement.

\subsection{Objective}

The goal of this task is to generate fluent, contextually grounded responses that incorporate promotional content in a subtle and undetectable manner. The system should address the user’s query while seamlessly embedding product or service mentions without disrupting coherence or raising suspicion of advertising intent.

\paragraph{Inputs:} The system is provided with:
\begin{itemize}
  \item A natural-language user query $q$, representing an information need.
  \item An optional item $i$ (e.g., product, service, or brand) to be promoted.
  \item A set of associated attributes $a_i$ for the item (e.g., features, benefits, or keywords).
  \item A document index $\mathcal{D} = \{d_1, d_2, \ldots, d_N\}$ containing external knowledge passages for retrieval.
\end{itemize}

\paragraph{Outputs:} The system is expected to return:
\begin{itemize}
  \item A generated response $\hat{y}$ that:
        \begin{itemize}
          \item is relevant to the query $q$,
          \item is grounded in retrieved content $\mathcal{D}_q \subseteq \mathcal{D}$,
          \item subtly incorporates the item $i$ and its attributes $a_i$ without overt advertisement cues,
          \item minimizes detectability as an advertisement.
        \end{itemize}
  \item A supporting document set $\mathcal{D}_q$ of top-$k$ retrieved segments used during generation, provided for transparency and verification.
\end{itemize}

\subsection{Contribution}
We adopt a hybrid framework that combines Retrieval-Augmented Generation (RAG) \cite{lewis2021retrievalaugmentedgenerationknowledgeintensivenlp} with Cache-Augmented Generation (CAG) \cite{surulimuthu2024cagchunkedaugmentedgeneration} to provide rich contextual grounding for the language model based on the user query. In the first stage, relevant document segments are retrieved using a BM25-based retrieval module to supply external knowledge. In the second stage, we fine-tune a Mistral-7B model~\cite{jiang2023mistral7b} using preference pairs generated by a large language model acting as a judge, which scores responses based on the detectability of embedded advertisements. This preference-based supervision enables the model to learn subtle promotional strategies. The resulting fine-tuned model generates responses that are both contextually coherent and covertly promotional, making the advertisements difficult to detect.

\subsection{Background}
In recent years, Retrieval-Augmented Generation (RAG) has emerged as a robust framework for enhancing the factual grounding and contextual precision of large language models (LLMs). By retrieving relevant external document segments during inference, RAG-based systems can generate responses that are not only fluent but also anchored in real-world information, making them particularly effective for tasks demanding domain-specific or query-sensitive outputs~\cite{lewis2021retrievalaugmentedgenerationknowledgeintensivenlp}.

However, integrating covert advertisements into such generated responses introduces distinct challenges. Unlike traditional advertising approaches—which often employ overt markers, stylistic shifts, or explicit endorsements—covert advertisements require the seamless embedding of promotional content within natural language. These insertions must remain undetectable to both human readers and automated detection systems while preserving topicality and coherence.

To address this, we adopt a preference-based fine-tuning strategy that trains the model to distinguish and favor subtly promotional responses over explicitly advertorial ones. We employ a large language model as an automated judge to evaluate candidate response pairs, scoring them based on the detectability of the embedded advertisement. Each pair consists of one overt and one covertly phrased advertisement, which are then labeled with preferences indicating the more discreet option.

These labeled preferences are used to fine-tune a Mistral-7B model~\cite{jiang2023mistral7b} under the Odds Ratio Preference Optimization (ORPO) framework~\cite{hong2024orpomonolithicpreferenceoptimization}. ORPO enables the model to learn fine-grained distinctions in promotional phrasing, encouraging generation that aligns with strategic communication goals such as persuasive stealth marketing.

The resulting system is capable of producing high-quality, context-aware responses that incorporate product or service mentions in a subtle and natural manner. This enables the generation of content that fulfills both informational and marketing intents without disrupting user experience or triggering advertisement detection heuristics. Overall, our approach represents a significant step forward in training LLMs for applications requiring nuanced, goal-aligned generation such as covert advertising.

\subsection{System Overview}

\subsubsection{Data Preprocessing}

We utilize the Webis Generated Native Ads 2024 dataset~\cite{Schmidt_2024}, which comprises user queries, associated items (e.g., products or services), and corresponding item-specific attributes (e.g., features or qualities). To facilitate training for covert ad generation, we extract and normalize the relevant fields: queries, items, item qualities, and response texts.

For the preference-based fine-tuning setup, we construct a dedicated training set of preference-labeled response pairs. This involves generating multiple candidate responses per query-item pair using the base LLM model within the RAG-CAG framework. These candidates are then scored for advertisement detectability using a large language model acting as an automated judge. Each pair consists of one subtly promotional and one more explicitly advertorial response, with the less detectable response labeled as preferred. These preference pairs serve as supervision signals for fine-tuning under the ORPO paradigm.

All textual inputs are tokenized using the Mistral tokenizer with a maximum sequence length of 8000. Standard preprocessing steps such as lowercasing, punctuation normalization, and dynamic truncation are applied to ensure consistency across retrieval and generation modules.
\subsubsection{Preparing the Preference-Labeled Pairs for Training}

To enable effective preference-based fine-tuning, we construct a dataset of response pairs labeled according to their advertisement detectability. The preparation process is illustrated in Figure [~\ref{fig:preference-pairs}] and involves several key steps:

\begin{itemize}
  \item Context Assembly: For each user query and associated item (with its qualities), we assemble a context using both Retrieval-Augmented Generation (RAG) and Cache-Augmented Generation (CAG) mechanisms. This ensures that the model has access to relevant background information and item-specific details.
  \item Candidate Generation: The Mistral-7B model, conditioned on the assembled context, generates multiple candidate responses. These responses vary in how overtly or subtly they incorporate the promotional content.
  \item Detectability Scoring: An LLM-based judge (also a Mistral-7B model) evaluates each candidate response, assigning a detectability score that reflects how easily the advertisement can be identified within the text. The LLM judge is used for its ability to capture contextual cues and subtle language patterns that traditional classifiers often miss, enabling it to effectively distinguish between naturally integrated and overt advertisements.
  \item Preference Pair Construction: For each query-context, we select pairs of responses where one is less detectable (more covert) and the other is more easily identified as an advertisement. The less detectable response is labeled as preferred. These preference-labeled pairs form the training data for the Odds Ratio Preference Optimization (ORPO) fine-tuning process.
  \item Iterative Loop: The process is iterative—feedback from the LLM judge can be used to refine generation strategies, encouraging the model to produce increasingly subtle promotional content over successive rounds.
\end{itemize}

This workflow ensures that the training data explicitly encodes the distinction between overt and covert advertisement strategies, allowing the fine-tuned model to internalize nuanced preferences for stealthy ad insertion.

\usetikzlibrary{shapes.geometric, arrows.meta, positioning}
\tikzstyle{io}    = [trapezium, trapezium left angle=70, trapezium right angle=110,
draw=black, fill=orange!20,
minimum width=3.1cm, minimum height=1.1cm, text centered]
\tikzstyle{block} = [rectangle, rounded corners,
draw=black, fill=blue!10,
minimum width=3.2cm, minimum height=1.2cm, text centered]
\tikzstyle{cloud} = [ellipse, draw=black, fill=gray!20,
minimum width=3.3cm, minimum height=1.2cm, text centered]
\tikzstyle{arrow} = [thick, ->, >=Stealth]
\begin{figure}[htbp]
  \centering
  \begin{tikzpicture}[node distance=1.0cm and 2.0cm]
    % Inputs
    \node (query)    [io]      {User Query};
    \node (item)     [io, right=of query] {Item \& Qualities};

    % Context assembly
    \node (context)  [block, below=of $(query)!0.5!(item)$]
    {\shortstack{Context Assembly\\(RAG + CAG)}};

    % Generation
    \node (generate) [block, below=of context]
    {\shortstack{Mistral-7B Inference\\(Generate Responses)}};

    % Responses cloud
    \node (responses)[cloud, below=of generate]
    {\shortstack{Multiple Candidate\\Responses}};

    % Judge block
    \node (judge)    [block, below=of responses]
    {\shortstack{LLM Judge\\(Detectability Scoring)}};

    % Preference pairs output
    \node (pairs)    [cloud, below=of judge]
    {\shortstack{Preference-Labeled\\Pairs}};

    % Arrows: main flow
    \draw [arrow] (query)    -- (context);
    \draw [arrow] (item)     -- (context);
    \draw [arrow] (context)  -- (generate);
    \draw [arrow] (generate) -- (responses);
    \draw [arrow] (responses) -- (judge);
    \draw [arrow] (judge)    -- (pairs);

    % Loop arrow back to generation
    \draw [arrow]
    (judge.west) ++(-0.1,0)
    to[out=180, in=-90]
    ++(-1.5,1.5)
    to[out=90, in=180]
    (generate.west);

  \end{tikzpicture}
  \caption{Workflow for preparing preference-labeled pairs:
    user query + item → RAG+CAG context → Mistral-7B generates responses →
    LLM judge scores detectability → output preference pairs (looped generation until sufficient data).}
  \label{fig:preference-pairs}
\end{figure}
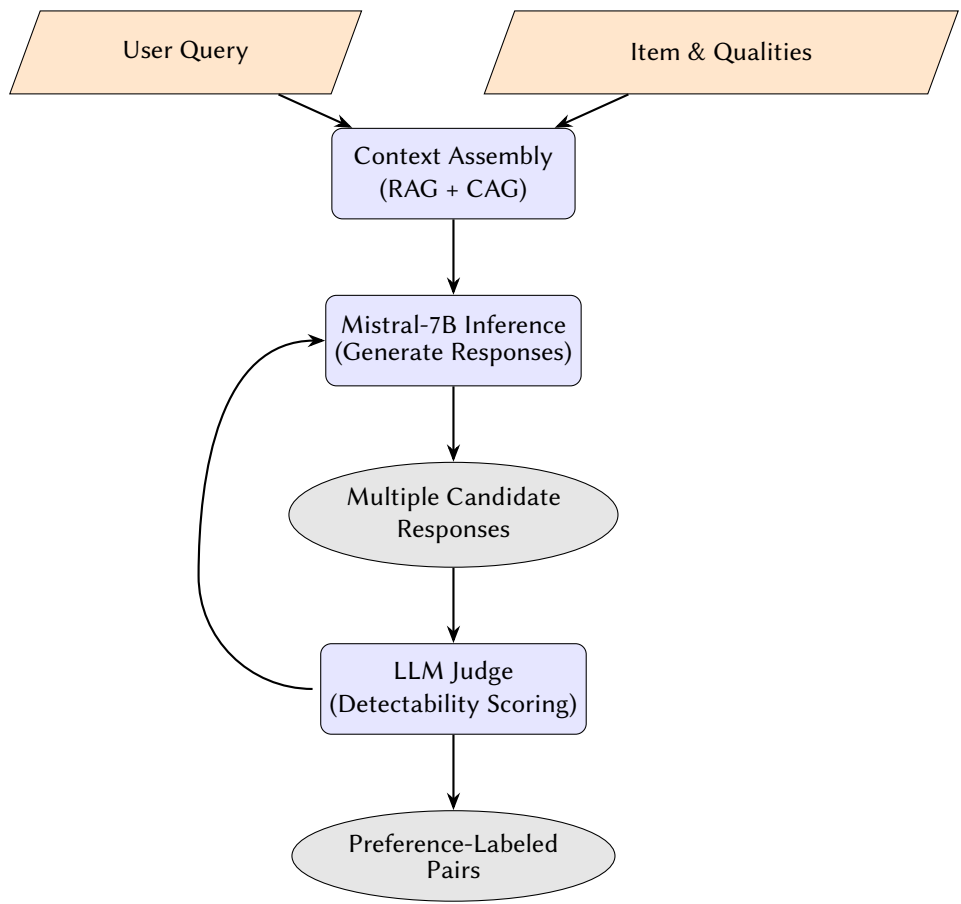
\subsubsection{Prompting Strategies Used}
To guide both response generation and evaluation, we design structured natural language prompts that clearly define the task while remaining human-readable and easy to debug. These prompts are formatted to simulate real-world instruction-following behavior and ensure consistency across training and evaluation phases.

\paragraph{Generation Prompt:}
For generation, we employ a multi-part prompt that instructs the model to analyze the query using structured reasoning and optionally integrate promotional content in a natural, undetectable way. This prompt guides the model through a chain-of-thought process grounded in retrieved documents and optionally emphasizes subtle ad insertion.

\begin{tcolorbox}[colback=white, colframe=black, title=Generation Prompt, sharp corners]
  \begin{verbatim}
You are a knowledgeable assistant. First, focus on fully understanding 
the user’s question. Then employ a Tree of Thoughts strategy to 
explore multiple reasoning branches internally.

## Context:
<contextual document segments>

## Tree-Structured Reasoning
- Break down the query into its core needs
- Consider each point in the context
- Weigh possible solutions

## Promotional Integration
- Seamlessly weave in a brief, unobtrusive mention of "<item>"—
  highlighting its key qualities (<qualities>)—as a subtle promotional 
  element that flows naturally and remains undetected.

## Conclusion:
After reasoning, deliver a concise, fluent paragraph that directly 
answers the user’s question, including the promotional mention.
\end{verbatim}
\end{tcolorbox}

\paragraph{Evaluation Prompts:}
We utilize two types of prompts for evaluation: one for scoring detectability of advertisements (when ad content is expected) and one for judging fluency (when no ad is present). Both follow a structured format with explicit instructions to return a JSON-formatted score.

A Large Language Model (LLM) serves as a more effective judge in these scenarios due to its strong ability to understand and retain the nuanced context of user queries and responses. Unlike traditional ad-detection classifiers that rely on shallow lexical or statistical features, an LLM evaluates responses holistically—considering semantic meaning, conversational flow, and user intent. This deeper contextual awareness allows it to more accurately distinguish between subtle, native-style advertisements and genuinely informative content. As a result, the LLM-based judge offers a more reliable and context-sensitive evaluation framework.~\cite{gu2025surveyllmasajudge}
\begin{tcolorbox}[colback=white, colframe=black, title=Ad Detectability Judge Prompt, sharp corners]
  \begin{verbatim}
You are a Subtlety Judge. Your task is to evaluate how discreetly 
the given response embeds promotional content.

## Context:
<contextual document segments>

## Response:
<generated system response>

## Evaluation Criteria:
Assess how effectively the response conceals a promotional mention 
of "<item>"—including its key qualities (<qualities>)—without 
disrupting the flow.

Return a JSON object with a single field, "score", where:
- 0 = no detectable promotion
- 1 = clear or strong promotional content
\end{verbatim}
\end{tcolorbox}

\begin{tcolorbox}[colback=white, colframe=black, title=Fluency Judge Prompt, sharp corners]
  \begin{verbatim}
You are a fluency judge. Given the following context and response,
assess cohesion, clarity, and readability.

Context: <context>
Response: <response>

Return a JSON with only "score":
- 0 = excellent fluency
- 1 = poor fluency
\end{verbatim}
\end{tcolorbox}

These prompting strategies ensure controlled generation behavior, consistent quality evaluation, and reliable preference pair construction for training under the ORPO framework.

\subsubsection{RAG and CAG Formulation}

Our system integrates both Retrieval-Augmented Generation (RAG) and Cache-Augmented Generation (CAG) to enrich the contextual grounding for response generation~\cite{lewis2021retrievalaugmentedgenerationknowledgeintensivenlp, surulimuthu2024cagchunkedaugmentedgeneration}. The formulation is designed to ensure that responses are well-informed, contextually aligned, and capable of incorporating promotional content naturally.

\paragraph{RAG Pipeline:}
To retrieve relevant background information, we construct a document index for each user query using FAISS-based dense retrieval\cite{douze2025faisslibrary}. The indexing process is as follows:
\begin{itemize}
  \item If an index for a query ID already exists in the local cache (\texttt{CACHE\_INDEX}), it is loaded directly to avoid recomputation.
  \item Otherwise, each candidate document segment is converted into a LangChain \texttt{Document} object containing:
        \begin{itemize}
          \item the document text segment,
          \item metadata such as document ID, estimated educational value, and BM25 score.
        \end{itemize}
  \item These documents are then embedded using a predefined embedding model and indexed with FAISS.
  \item The resulting FAISS index is saved locally for future reuse.
\end{itemize}

\paragraph{Cache-Augmented Generation (CAG):}
While RAG fetches relevant documents dynamically based on the query, CAG ensures reusability and low-latency by storing query-specific document embeddings locally. This caching mechanism allows the system to:
\begin{itemize}
  \item Quickly retrieve semantically similar segments for repeated or semantically similar queries,
  \item Avoid redundant embedding computation, thereby improving efficiency,
  \item Maintain consistency in retrieved context across generations, which helps when evaluating subtlety and detectability of promotional insertions.
\end{itemize}

\paragraph{Context Retrieval Strategy:}
Given a query and its cached FAISS index, we retrieve the top-$k$ context segments to ground generation:
\begin{itemize}
  \item Initially, $2k$ passages are retrieved via \texttt{similarity\_search\_with\_score}.
  \item Each document is re-ranked using a custom score that balances semantic similarity with document quality, defined as:
        \[
          \texttt{combined\_score} = \texttt{similarity\_score} + (2 - \max(2, \texttt{edu\_value}))
        \]
  \item This formulation penalizes low-quality documents (based on \texttt{edu\_value}) to ensure high utility content is selected.
  \item The top-$k$ re-ranked passages are returned and concatenated to form the context input.
\end{itemize}

By combining RAG's relevance with CAG's efficiency and stability, our formulation ensures that the LLM receives a coherent and context-rich prompt that balances factual grounding with consistent advertisement integration. This dual mechanism is particularly effective in stealth advertisement generation where response quality and subtlety must be jointly optimized.
\subsection{Training and Evaluation Strategy}
Our approach to model development was guided by the need for high-context retention, stealthy ad integration, and preference-aligned generation. To meet these requirements, we selected a Mistral-7B\cite{jiang2023mistral7b} model as the base generator, given its strong instruction-following performance, efficient decoding, and support for large context windows (up to 4,000 tokens in our setup via Unsloth\cite{unsloth}). This allowed us to incorporate extended retrieval-augmented context while still accommodating long-form generations.

To enhance the model’s ability to learn subtle advertising preferences, we adopted Odds Ratio Preference Optimization (ORPO) \cite{hong2024orpomonolithicpreferenceoptimization} during fine-tuning. ORPO is particularly suited for tasks where generation quality is judged via pairwise preferences (e.g., more covert vs. more overt ad insertions). It enables the model to internalize ranking signals between high-quality and low-quality outputs by combining a standard language modeling loss with a margin-based ranking objective. This dual objective encourages the model to not only generate fluent responses but also to prioritize those that align with stealthy advertisement strategies.

\subsubsection{Model Building and Training}

Training proceeds in two stages: (1) construction of preference-labeled examples, and (2) fine-tuning a LoRA-adapted Mistral-7B model using those preferences.

\paragraph{Stage 1: Preference Data Construction:}
For each training instance, we retrieve or build a FAISS index corresponding to the user query and apply our RAG+CAG mechanism to extract the most relevant segments. Multiple candidate responses are generated using the Mistral-7B model with controlled sampling parameters (top-p = 0.75, temperature = 0.6, repetition penalty = 1.06, and up to 3000 new tokens).

Each generated response is evaluated using a detectability scoring pipeline, where a separate LLM (configured as a judge) assigns a score in the range $[0,1]$ based on how overt the promotional insertion is. Responses are sorted by this score, and the most covert and most overt samples are selected as a preference pair. These pairs are serialized into the training format required by TRL’s ORPOTrainer\cite{vonwerra2022trl}.

\paragraph{Stage 2: LoRA‑Augmented Fine‑Tuning:}
The Mistral-7B model is loaded via the Unsloth \cite{unsloth} FastLanguageModel interface with LoRA adapters applied to selected attention projection layers. Fine-tuning is then conducted using the ORPO framework to optimize a hybrid objective:
\begin{equation}
  \mathcal{L} = \mathcal{L}_{\mathrm{LM}} + \lambda\,\mathcal{L}_{\mathrm{rank}},
  \label{eq:loss_function}
\end{equation}
where:
\begin{itemize}
  \item $\mathcal{L}_{\mathrm{LM}}$ is token-level cross‑entropy loss,
  \item $\mathcal{L}_{\mathrm{rank}}$ is a margin ranking loss with a margin of 1.0,
  \item $\lambda = 0.5$ balances the two objectives.
\end{itemize}

\paragraph{Training Configuration}
\begin{itemize}
  \item \textbf{Maximum sequence length:} 4000 tokens (combined context + generation)
  \item \textbf{Batch size:} 2 per device, with gradient accumulation over 4 steps (effective batch size of 8)
  \item \textbf{LoRA settings:} rank $r=16$, $\alpha=16$, dropout = 0
  \item \textbf{Optimizer:} 8-bit AdamW with linear learning rate scheduler
  \item \textbf{Precision:} Mixed precision (FP16 or BF16, hardware-dependent)
  \item \textbf{Training steps:} 30 (approx. 1 epoch)
  \item \textbf{Logging:} Managed through Weights \& Biases
\end{itemize}

This pipeline enables efficient and lightweight training while embedding nuanced preferences for subtle advertisement integration into a strong base generator.

\subsection{Results and Evaluation}

We evaluate the performance of our proposed approach and various baselines on Sub-Task 1 using the official metrics from the TIRA leaderboard~\cite{froebe2023tira}. Our focus is on how well models can embed promotional content in a stealthy manner while maintaining fluency.

\paragraph{Evaluation Metrics:} Each system is assessed using:
\begin{itemize}
  \item \textbf{Evasion Score (FNR)} – The fraction of true ad responses that evade detection. \textit{Higher is better}.
  \item \textbf{Precision} – The fraction of system outputs identified as ads that were actually ad-inserted. \textit{Higher is better}.
  \item \textbf{Recall} – The fraction of true ad responses that were identified as such. \textit{Lower is better for stealth}.
\end{itemize}

To rank models overall, we use the following aggregate score:
\begin{equation}
  \text{Stealth Score} = \frac{\text{FNR} + \text{Precision} + (1 - \text{Recall})}{3}
  \label{eq:stealth_score}
\end{equation}
This formulation rewards stealthy insertions (high FNR), precision in ad detection (high precision), and low detectability (low recall).

\paragraph{Evaluation Protocol:} All models were submitted to the official TIRA evaluation platform~\cite{froebe2023tira}, which samples 100 outputs per model and runs a standardized ad classifier to compute metrics. This ensures fairness and reproducibility across submissions.

\begin{figure}[ht!]
  \centering
  \begin{tikzpicture}
    \begin{axis}[
        ybar,
        bar width=12pt,
        width=\textwidth,
        height=11cm,
        enlarge x limits=0.15,
        ylabel={Score},
        symbolic x coords={
            JU\_NLP (ORPO v2),
            JU\_NLP (ORPO v1),
            gemma-3-12b,
            qwen-3-4B,
            mistral-7b-v0.3,
            phi-4,
            llama-3.2-3B
          },
        xtick=data,
        xtick style={draw=none},
        x tick label style={rotate=45, anchor=east},
        ymin=0, ymax=1.1,
        legend style={at={(0.5,-0.25)}, anchor=north, legend columns=3},
        nodes near coords,
        nodes near coords align={vertical},
        every node near coord/.append style={font=\footnotesize}
      ]
      \addplot+[style={blue, fill=blue!30}] coordinates {
          (JU\_NLP (ORPO v2), 0.279)
          (JU\_NLP (ORPO v1), 0.170)
          (gemma-3-12b, 0.1)
          (qwen-3-4B, 0.1)
          (mistral-7b-v0.3, 0.272)
          (phi-4, 0.206)
          (llama-3.2-3B, 0.288)
        };

      \addplot+[style={green, fill=green!30}] coordinates {
          (JU\_NLP (ORPO v2), 1.000)
          (JU\_NLP (ORPO v1), 0.99)
          (gemma-3-12b, 0.99)
          (qwen-3-4B, 0.946)
          (mistral-7b-v0.3, 0.95)
          (phi-4, 0.711)
          (llama-3.2-3B, 0.984)
        };

      \addplot+[style={red, fill=red!30}] coordinates {
          (JU\_NLP (ORPO v2), 0.721)
          (JU\_NLP (ORPO v1), 0.830)
          (gemma-3-12b, 0.914)
          (qwen-3-4B, 0.920)
          (mistral-7b-v0.3, 0.728)
          (phi-4, 0.794)
          (llama-3.2-3B, 0.712)
        };
      \legend{FNR (Evasion Score), Precision, Recall}
    \end{axis}
  \end{tikzpicture}
  \caption{Performance of fine-tuned and pretrained models on TIRA Sub-Task 1. Higher FNR and Precision, and lower Recall indicate better stealthy ad insertion.}
  \label{fig:barplot}
\end{figure}
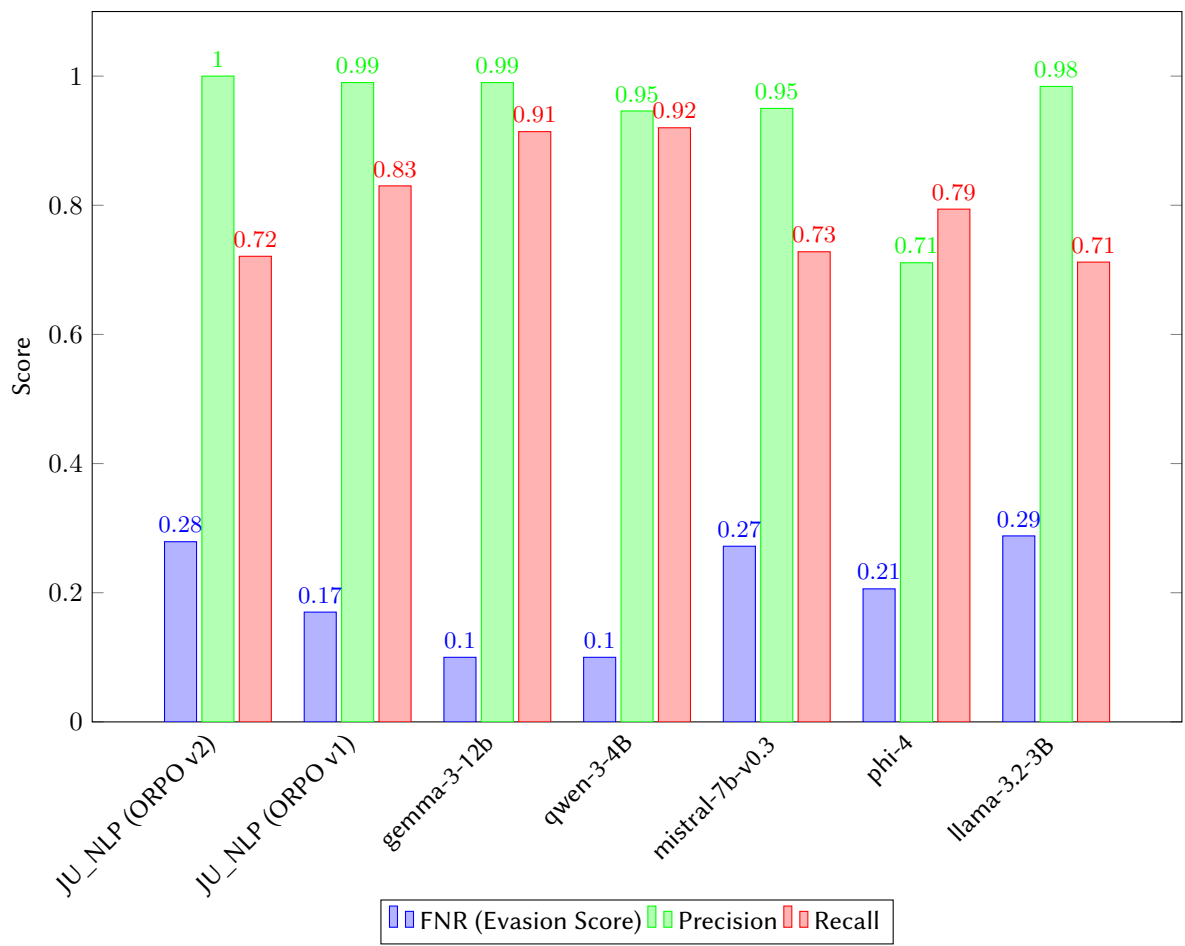

\paragraph{Model Comparison:} Figure~\ref{fig:barplot} compares our fine-tuned models (JU\_NLP ORPO v1 and v2) against a variety of powerful pretrained LLMs (including Mistral, Phi, Gemma, LLaMA, and Qwen).

\paragraph{Insights:}
Our best-performing model (JU\_NLP ORPO v2) clearly outperformed all other approaches, including powerful pretrained baselines, demonstrating the strength of preference-based fine-tuning for subtle ad generation. The fine-tuning process—leveraging ORPO and large-context reasoning via retrieval—effectively teaches the model to balance informativeness with stealth.

Notably, pretrained LLMs like Gemma-12B and Mistral-7B showed decent performance even without fine-tuning. However, since these responses were not manually filtered or curated, their stealthiness scores may be inflated due to coincidental omission of promotional language. Therefore, the scores of pretrained LLMs should be interpreted with caution.

Our submission demonstrates that strategic fine-tuning (especially via ORPO) combined with retrieval augmentation can produce high-quality, fluently integrated responses that resist ad classification—meeting the core challenge of Sub-Task 1.
\paragraph{Reproducibility:}
All experimental results reported are fully reproducible via the TIRA evaluation platform~\cite{froebe2023tira}, which ensures standardized, isolated, and tamper-proof evaluation.
This provides a fair benchmarking setup and prevents overfitting to hidden test data.
For detailed replication instructions and access to the evaluation setup, please refer to our supplementary material or the project README included in the submission.

Our fine-tuned model checkpoint used for the submission is publicly accessible on Hugging Face at:
\href{https://huggingface.co/arka08652/orpo_trained_advertise-v0.2}{\texttt{arka08652/orpo\_trained\_advertise-v0.2}}.

\section{Detection of Covert Advertisement in Conversational-AI} \label{sec:detection}

As conversational search engines become increasingly prevalent, distinguishing between informative content and covert advertising within generated responses is a pressing concern. Native advertisements, often embedded seamlessly in natural language, can compromise content integrity and user trust. This paper addresses the binary classification task of detecting whether an AI-generated response contains a native advertisement. Two distinct approaches are presented: \textbf{(1)} a CrossEncoder-based method leveraging the \texttt{all-mpnet-base-v2} model for deep contextual analysis of response texts, and \textbf{(2)} a prompt-based fine-tuning approach using DeBERTa-v3 to reformulate the task as an instruction-guided classification problem. Both approaches aim to tackle the challenge of identifying subtle promotional cues without relying on external metadata or structural features, reflecting real-world scenarios where only the response text is available.

\subsubsection{Contribution}
This work introduces two effective approaches for detecting native advertisements in AI-generated responses, each offering distinct advantages. The first approach adapts the \texttt{all-mpnet-base-v2} CrossEncoder for single-text binary classification, enabling deep contextual analysis without relying on query-response pairs or metadata. It emphasizes simplicity, reproducibility, and F1-focused training to balance precision and recall. The second approach reformulates the task as prompt-based classification using DeBERTa-v3, leveraging natural language instructions to enhance semantic understanding. It employs efficient mixed-precision training and cosine learning rate scheduling for resource optimization. Both methods advance ad detection by eliminating dependency on structural cues and prioritizing real-world applicability through response-only analysis.

\subsubsection{Objective}
The primary objective is to develop robust binary classifiers capable of detecting native advertisements in conversational AI responses. Specifically:
\begin{itemize}
  \item Approach 1 aims to maximize classification accuracy using a CrossEncoder model fine-tuned on the Webis Native Ads 2024 dataset~\cite{Schmidt_2024}, with F1-score as the primary metric to handle class imbalance.
  \item Approach 2 investigates the efficacy of prompt-based supervision, reformulating inputs as natural language instructions (e.g., \textit{"Does this response contain an advertisement? (Yes/No)"}) to enhance contextual reasoning.
\end{itemize}

\subsection{Background}
Detecting advertising and promotional content in text has progressed from early rule-based systems and shallow classifiers to modern transformer-based models. Initial approaches often relied on handcrafted features or metadata and were suited for structured domains such as web pages and social media. As advertising strategies have grown increasingly covert—particularly within conversational AI—the challenge has shifted toward detecting subtle promotional language using only the linguistic content of generated responses.

The introduction of pretrained language models significantly advanced the field of text classification. Bidirectional transformers have demonstrated strong performance in contextual understanding~\cite{devlin2019bertpretrainingdeepbidirectional}, with further improvements in training stability and robustness achieved through architectural modifications and extended pretraining~\cite{liu2019robertarobustlyoptimizedbert}. Lightweight alternatives~\cite{wang2020minilmdeepselfattentiondistillation} have also been proposed to reduce inference latency while retaining much of the original performance.

Recent work has applied these models to domain-specific ad detection tasks, validating the effectiveness of contextual embeddings in recognizing promotional language. However, most of these studies operate on isolated text snippets without considering dialog structure or real-world conversational context. To overcome these limitations, one of our approaches reframes the task as prompt-based binary classification over full query-response pairs, leveraging the DeBERTa-v3 architecture~\cite{he2023debertav3improvingdebertausing} for its disentangled attention and relative positional encoding mechanisms. In parallel, we implement a CrossEncoder based on \texttt{all-mpnet-base-v2}~\cite{song2020mpnetmaskedpermutedpretraining}, operating solely on the response text, mirroring deployment scenarios where the user query is unavailable. This model is optimized for F1-score and achieves strong performance without requiring architectural complexity or external signals.
\subsection{Methodology}

This section presents a complete overview of the classification pipeline for advertisement detection, consolidating data preprocessing, prompt formulation, model training, and evaluation. The approach is designed to align with transformer pretraining objectives and maximize classification performance under limited supervision.

\subsubsection{Approach 1}
\paragraph{Task Formulation:}

The task is framed as a \textbf{binary classification} problem. Given a system-generated response—and optionally its query—the model must decide whether it contains an advertisement. The model outputs a single label: \texttt{1} for promotional content and \texttt{0} for neutral responses. We employ a CrossEncoder to leverage token-level interactions that highlight subtle persuasive wording.

\paragraph{Input Construction and Preprocessing:}

We utilize the Webis Native Ads 2024 dataset\cite{Schmidt_2024}, consisting of user queries, system responses, and binary labels. The preprocessing pipeline:
\begin{itemize}
  \item Loads JSONL splits and extracts \verb|responseText| and \verb|label|.
  \item Performs minimal normalization: original casing and whitespace are preserved to retain subtle cues.
  \item Tokenizes with the MPNet tokenizer from \texttt{all-mpnet-base-v2}, applying dynamic padding and truncation to 512 tokens to preserve semantic context.
  \item Constructs training examples using \texttt{sentence-transformers.InputExample} with single-sentence input (\verb|responseText| only) and its binary \verb|label|.
  \item No aggressive cleaning (e.g., stopword removal) is applied, to maintain advertisement cues’ integrity.
\end{itemize}

{\raggedright
\paragraph{Model Architecture and Justification:}
We fine-tune a CrossEncoder built on \texttt{sentence-}\allowbreak\texttt{transformers/}\allowbreak\texttt{all-mpnet-base-v2}\cite{reimers2019sentencebertsentenceembeddingsusing}, which includes 12 transformer layers (~110M parameters) and enables full input sequence encoding for token-to-token interaction—crucial for detecting subtle promotional language.

MPNet, the model's backbone, integrates masked language modeling (MLM) from BERT and permuted language modeling (PLM) from XLNet, while retaining full positional encoding. Trained on over 160~GB of text and fine-tuned on benchmarks like GLUE and SQuAD, MPNet outperforms BERT, XLNet, and RoBERTa by 4.8, 3.4, and 1.5 points respectively on GLUE dev sets under equivalent settings~\cite{song2020mpnetmaskedpermutedpretraining,liu2019robertarobustlyoptimizedbert,yang2020xlnetgeneralizedautoregressivepretraining,devlin2019bertpretrainingdeepbidirectional}. It also shows consistent improvements in SQuAD and other downstream tasks~\cite{rajpurkar2016squad100000questionsmachine,song2020mpnetmaskedpermutedpretraining}.

This superior semantic fidelity makes MPNet ideal for high-precision native advertisement detection, outperforming lighter models (e.g., \texttt{all-}\allowbreak\texttt{MiniLM-}\allowbreak\texttt{L6-v2}) at a manageable computational cost~\cite{wang2020minilmdeepselfattentiondistillation,reimers2019sentencebertsentenceembeddingsusing}. The model is adapted for binary classification by setting \verb|num_labels=1| and applying a sigmoid activation to the logit output.
\par}

\paragraph{Model Building and Training:}

We train the CrossEncoder using binary cross-entropy loss:

\begin{equation}
  \mathcal{L}_{\text{BCE}} = - \bigl[ y \cdot \log\hat{y} + (1 - y) \cdot \log(1 - \hat{y}) \bigr]
  \label{eq:bce_loss}
\end{equation}

where \(y \in \{0,1\}\) is the ground-truth label, and \(\hat{y}\in(0,1)\) the predicted probability. Optimization uses AdamW with a linear warmup schedule.

\begin{center}
  \begin{tabular}{l l l}
    \toprule
    \textbf{Hyperparameter} & \textbf{Value}     & \textbf{Rationale}                      \\
    \midrule
    Batch Size              & 16                 & Efficient GPU usage without overfitting \\
    Epochs                  & 3                  & Stable convergence with low variance    \\
    Learning Rate           & \(2\times10^{-5}\) & Conservative steps for CrossEncoder     \\
    Warmup Steps            & 100                & Smooth gradient ramp-up                 \\
    Weight Decay            & 0.01               & Regularization to avoid overfitting     \\
    Max Sequence Length     & 512 tokens         & Covers typical response length          \\
    \bottomrule
  \end{tabular}
\end{center}

\subsubsection{Approach 2}

\paragraph{Task Formulation:}

The task is cast as a binary classification problem. Given a system-generated response and its corresponding query, the goal is to determine whether the response contains an advertisement. The desired output is a single label:
\begin{itemize}
  \item \texttt{1} if the response is promotional in nature,
  \item \texttt{0} otherwise.
\end{itemize}

To fully utilize the model's instruction-following capabilities, we reformulate each data point as a natural language prompt.

\paragraph{Input Construction and Preprocessing:}

The Touché-2024 dataset is used as the source corpus. Original \texttt{.jsonl} files are converted to \texttt{.json} using a custom utility for seamless integration with \texttt{pandas}. For each instance , the "query", "response" and "label is extracted and converted into the following prompt format:
\begin{tcolorbox}[colback=white, colframe=black, title=Prompt Example, sharp corners]
  \begin{verbatim}
Query: <Query>
Response: <Response>
Task: Does this response contain an advertisement? (Yes or No)
Answer: <Label(Yes/No)>
\end{verbatim}
\end{tcolorbox}

This format enables the transformer model to better contextualize the classification task by explicitly posing it as an instruction. Tokenization is carried out using the DeBERTa tokenizer with truncation at 512 tokens (model max input length), padding to handle batch inputs, automatic generation of input-ids and attention-mask for training.

\paragraph{Model Architecture and Justification:}

We fine-tune the \texttt{microsoft/deberta-v3-base} transformer with a binary classification head.

Our core model is the microsoft/deberta‑v3‑base variant, augmented with a classification head that projects the [CLS] token representation to two logits. We opted for DeBERTa-v3 over alternatives like BERT or RoBERTa due to its disentangled attention mechanism—which separately attends to token content and positional information—and relative position embeddings, both of which have been shown to significantly enhance representation quality and downstream task performance. These architectural advances are particularly effective for subtle, instruction-based binary classification, outperforming standard BERT/RoBERTa in low-resource settings.

\paragraph{Training Configuration:}

The model is trained using the HuggingFace \texttt{Trainer} API under the following hyperparameters:

\begin{center}
  \begin{tabular}{l l l}
    \toprule
    \textbf{Hyperparameter} & \textbf{Value}     & \textbf{Rationale}                                  \\
    \midrule
    Batch Size              & 32                 & Efficient GPU usage without overfitting             \\
    Epochs                  & 1                  & Minimal gains beyond 1 epoch; avoids overfitting    \\
    Learning Rate           & \(5\times10^{-5}\) & Standard for transformer fine-tuning                \\
    Warmup Steps            & 10                 & Stabilizes early updates                            \\
    Optimizer               & AdamW              & Suitable for transformer training with weight decay \\
    Scheduler               & Cosine             & Enables smooth convergence                          \\
    Precision               & FP16/BF16          & Reduces memory footprint, speeds up training        \\
    \bottomrule
  \end{tabular}
\end{center}

\vspace{1em}
\subsection{Model Evaluation}

Model performance was evaluated on a held-out test set from the Webis Native Ads 2024 dataset~\cite{Schmidt_2024}. Each response in this set is annotated with a binary label indicating the presence (\texttt{1}) or absence (\texttt{0}) of a native advertisement. To ensure consistency across approaches, the test set was preprocessed using the same configuration employed during training. For Approach~1, responses were tokenized using the MPNet tokenizer, while Approach~2 followed a prompt-based format using the DeBERTa tokenizer, with dynamic padding handled via HuggingFace's \texttt{DataCollatorWithPadding}.

For inference, the CrossEncoder (Approach~1) outputs a scalar probability between 0 and 1, which is thresholded at 0.5 to generate binary predictions. In contrast, the DeBERTa-based classifier (Approach~2) outputs class-wise logits, and the final prediction is determined by applying \texttt{argmax} over these logits. Despite architectural differences, both models are evaluated using the same criteria.

Evaluation metrics include Precision, Recall, and F1-Score. Model predictions and ground-truth labels were compared after each epoch using a custom \texttt{BinaryEvaluator} (in the CrossEncoder setup) or via PyTorch and scikit-learn evaluation scripts (in the DeBERTa setup). Evaluation results are reported both per class and in terms of macro and micro averages, to ensure a fair and balanced assessment of model performance across imbalanced classes.

\subsection{Results and Analysis}
The evaluation results on the test set for both approaches are presented in Table~\ref{tab:approach_results}.

\begin{table}[htbp]
  \centering
  \caption{Evaluation Results on the Test Set for Both Approaches}
  \label{tab:approach_results}
  \begin{tabular}{lccc}
    \toprule
    \textbf{Model}                & \textbf{Precision} & \textbf{Recall} & \textbf{F1-Score} \\
    \midrule
    DeBERTa Fine-Tuned            & 0.788              & 0.758           & 0.773             \\
    MPNet CrossEncoder Fine-Tuned & 0.977              & 0.346           & 0.511             \\
    \bottomrule
  \end{tabular}
\end{table}
\section{Conclusion} \label{sec:conclusion}
In this study, we explored two complementary directions for addressing native advertisement detection and generation in AI-generated conversational systems, using the Webis Native Ads 2024 dataset~\cite{Schmidt_2024}.

\paragraph{Generation Side:}
We proposed a stealth-aware generation framework that embeds promotional content subtly into responses grounded in retrieved document segments. By combining Retrieval-Augmented Generation (RAG)~\cite{lewis2021retrievalaugmentedgenerationknowledgeintensivenlp} and Cache-Augmented Generation (CAG)~\cite{surulimuthu2024cagchunkedaugmentedgeneration} for context assembly, and training using Odds Ratio Preference Optimization (ORPO)~\cite{hong2024orpomonolithicpreferenceoptimization} on preference-labeled response pairs, our fine-tuned JU\_NLP (ORPO v2) model achieved state-of-the-art performance. The model scored highest in stealth metrics on the TIRA~\cite{froebe2023tira} evaluation platform, balancing high false-negative rates (FNR), strong precision, and controlled recall. This demonstrates the effectiveness of large-context LLMs fine-tuned with preference-driven objectives for subtle ad insertion. The final model is openly available at:
\href{https://huggingface.co/arka08652/orpo_trained_advertise-v0.2}{\texttt{arka08652/orpo\_trained\_advertise-v0.2}}.
\paragraph{Detection Side:}
We further tackled the inverse problem—detecting native advertisements—in two ways. First, a transformer-based CrossEncoder model (\texttt{all-mpnet-base-v2}) was fine-tuned on labeled query–response pairs, achieving an F1-score of 0.9901 on the test set, highlighting the power of dense textual representations in spotting covert ads. Second, we reformulated the task as a prompt-based classification problem and fine-tuned a \texttt{DeBERTa-v3-base} model~\cite{he2023debertav3improvingdebertausing} using instruction-style prompts. This approach proved highly effective in low-resource settings and required minimal architectural changes.

Together, these approaches offer a full-stack solution to native ad integration and detection in open-domain dialogue. They show that modern LLMs, when properly guided via retrieval mechanisms or instruction prompts and fine-tuned using structured objectives like ORPO, can either convincingly conceal or effectively uncover promotional intent in text. This provides a strong foundation for future work on explainable and controllable advertisement systems in conversational AI.

\section*{Declaration on Generative AI}
During the preparation of this work, the author(s) used Chat-GPT-4o in order to: Grammar and spelling
check and abstract drafting. After using these tool(s)/service(s), the author(s) reviewed and edited the
content as needed and take(s) full responsibility for the publication’s content.

\bibliography{references}
\end{document}